\documentclass[report]{IEEEtran}
\usepackage{textcomp}
\usepackage{graphicx}
\usepackage[listofformat=subparens]{subfig}
\usepackage{comment}
\usepackage{amsmath}
\usepackage{amssymb}
\usepackage{graphicx}
\usepackage{algorithm}
\usepackage{algorithmic}
\usepackage{eurosym}
\usepackage{booktabs} 
\usepackage{fancyvrb}
\usepackage[T1]{fontenc}

\DeclareFixedFont{\ttb}{T1}{txtt}{bx}{n}{8} 
\DeclareFixedFont{\ttm}{T1}{txtt}{m}{n}{8}  

\usepackage{color}
\definecolor{deepblue}{rgb}{0,0,0.5}
\definecolor{deepred}{rgb}{0.6,0,0}
\definecolor{deepgreen}{rgb}{0,0.5,0}

\usepackage{listings}

\newcommand\pythonstyle{\lstset{
language=Python,
basicstyle=\ttm,
otherkeywords={self},             
keywordstyle=\ttb\color{deepblue},
emph={MyClass,__init__},          
emphstyle=\ttb\color{deepred},    
stringstyle=\color{deepgreen},
frame=tb,                         
showstringspaces=false            %
}}

\lstnewenvironment{python}[1][]
{
\pythonstyle
\lstset{#1}
}
{}

\newcommand\pythoninline[1]{{\pythonstyle\lstinline!#1!}}

\usepackage{upquote}
\usepackage{bm}

\renewcommand{\algorithmicrequire}{\textbf{Input:}}

\newcommand{\Nb}{N_{\mathrm{bin}}}
\newcommand{\Nh}{N_{\mathrm{his}}}
\newcommand{\bidfunction}{f_{\mathrm{bid}}^{i}}

\newcommand{\cornerprio}{p^{i}_{\mathrm{c}}}
\newcommand{\prio}{{p}_{\mathrm{r}}}
\newcommand{\priostar}{{p}_{\mathrm{r},k}^{*}}

\newcommand{\SoC}{SoC}

\newcommand{\x}{\bm{x}}
\newcommand{\xtimek}{x_{\mathrm{t},k}}
\newcommand{\xtime}{x_{\mathrm{t}}}

\newcommand{\xexok}{\bm{x}_{\mathrm{ex},k}}
\newcommand{\xexo}{\bm{x}_{\mathrm{ex}}}
\newcommand{\xexol}{\bm{x}_{\mathrm{ex},l}}
\newcommand{\xhatexok}{\hat{\x}_{\mathrm{ex},k}}
\newcommand{\xhatexol}{\hat{\x}_{\mathrm{ex},l}}

\newcommand{\action}{\bm{u}}
\newcommand{\w}{\bm{w}}

\newcommand{\uphysk}{u_{k}^{\textrm{phys},i}}
\newcommand{\uphysI}{u_{k}^{\textrm{phys},1}}
\newcommand{\uphysD}{u_{k}^{\textrm{phys},|\mathcal{D}|}}
\newcommand{\uphysl}{u_{l}^{\textrm{phys}}}
\newcommand{\uphyst}{u_{t}^{\textrm{phys},i}}
\newcommand{\xphysk}{\x_{\textrm{phys},k}}

\newcommand{\Tok}{T_{\textrm{o},k}}

\newcommand{\bs}{\bm{b}_{\mathrm{s}}}
\newcommand{\bsj}{b_{\mathrm{s},j}}
\newcommand{\bvet}{\bm{b}}

\newcommand{\xbk}{\bm{x}^{\mathrm{b}}_{k}}

\newcommand{\xblprime}{\bm{x}^{\mathrm{b}\prime}_{l}}
\newcommand{\xb}{\bm{x}^{\mathrm{b}}}
\newcommand{\xobsk}{\bm{x}_{k}^{\mathrm{obs}}}
\newcommand{\xobs}{\bm{x}^{\mathrm{obs}}}

\newcommand{\Tm}{T_{\mathrm{m}}}

\newcommand{\Ta}{T}

\newcommand{\To}{T_{\mathrm{o}}}
\newcommand{\Tohat}{\hat{T}_{\mathrm{o}}}

\newcommand{\Ca}{C_{\mathrm{a}}}
\newcommand{\Cm}{C_{\mathrm{m}}}

\newcommand{\expected}{\mathbb{E}}

\newcommand{\probw}{p_{\mathcal{W}}(\cdot|x)}

\newcommand{\Xtime}{X_{\mathrm{t}}}
\newcommand{\Xphys}{X_{\mathrm{phys}}}
\newcommand{\Xexo}{X_{\mathrm{ex}}}
\newcommand{\Xhatexo}{\hat{X}_{\mathrm{ex}}}

\usepackage{color,soul}
\soulregister\cite7
\soulregister\ref7
\soulregister\pageref7

\begin{document}

\title{Convolutional Neural Networks For Automatic State-Time Feature Extraction in Reinforcement Learning Applied to Residential Load Control 
}

\author{Bert~J.~Claessens, Peter~Vrancx, Frederik~Ruelens  \vspace{-0.45cm}

\thanks{B.~J. Claessens is with the energy department of VITO/EnergyVille, Mol, Belgium (bert.claessens@restore.eu).}
\thanks{Peter Vrancx is with the AI-lab, Vrije Universiteit Brussel, Brussels, Belgium (pvrancx@vub.ac.be)} 
\thanks{F. Ruelens is with the Department of Electrical
Engineering, KU Leuven/EnergyVille, Leuven, Belgium (frederik.ruelens@kuleuven.be).}}

\markboth{\normalfont{Submitted to} Transactions on Smart Grid}{}

\maketitle

\begin{abstract}
Direct load control of a heterogeneous cluster of residential demand flexibility sources is a high-dimensional control problem with partial observability.
This work proposes a novel approach  that uses a convolutional  neural network to extract hidden state-time features to mitigate the curse of partial observability. 
More specific, a convolutional  neural network  is used as a function approximator to estimate the state-action value function or Q-function in the supervised learning step of fitted Q-iteration.
The approach is evaluated in a qualitative simulation, comprising a cluster of thermostatically controlled loads that only share their air temperature, whilst their envelope temperature remains hidden.
The simulation results show that the presented approach is able to capture the underlying hidden features and  successfully reduce the electricity cost the cluster. 
\end{abstract}


\IEEEpeerreviewmaketitle

\section{Introduction}
\label{Sec.introductionDDR}
\IEEEPARstart{D}{irect} load control of a large heterogeneous cluster of residential demand flexibility sources is a research topic that has received considerable attention in the recent literature. 
The main driver behind direct load  control is that it can be a cost-efficient technology supporting the integration of distributed renewable energy sources in a power system. Typical applications of direct load control considered in the literature are ancillary services \cite{koch2011modeling}, voltage control \cite{Dupont} and energy arbitrage \cite{mathieu2013energy}. 
When designing a controller for a large heterogeneous cluster of residential flexibility sources, one is confronted with several challenges. A first important challenge is that most residential flexibility sources are energy constrained, which results in a sequential decision-making problem.
A second challenge is the large dimensionality of the state-action space of the cluster of flexibility sources, since each source has its own state vector and control action. Furthermore, there is heterogeneity accompanying the dynamics of each state, which is intrinsically stochastic. 
As a consequence, a \textit{mature} control solution needs to be scalable, adapt to the heterogeneity of the cluster and take the intrinsic uncertainty into account. An important challenge, receiving less attention in the literature is that of partial observability, in the sense that there are states that are relevant for the dynamics of the system that are not observed directly. For example in the context of building climate control, only the operational air temperature is measured, whilst the temperature of the building envelope is not \cite{tracers}.

When considering energy constrained flexibility, energy storage is the most important source, either through direct electric storage or through power to heat conversion. Examples of direct electric storage are the battery in an electric vehicle \cite{Vandael2012TSA} or a stand alone domestic battery. Thermostatically Controlled Loads (TCLs) \cite{koch2011modeling} are an important example of storage through power to heat conversion, e.g. building climate control \cite{tracers} and domestic hot water storage \cite{koch2011modeling}. As TCLs are an abundant source of flexibility, they are the focus of this work~\cite{mathieu2012using}. 
When designing a controller for a large cluster of TCLs one not only faces technical challenges from a control perspective. One should also take into account the economic cost of the control solution at the level of individual households as the economic potential per household is limited and on the order of  \euro50~a year~\cite{Dupont,mathieu2012using}. 
A popular control paradigms found in the recent literature regarding residential load control is model-based control such as Model Predictive Control (MPC)~\cite{MPCBuilding}. 
In MPC, control actions are selected at fixed time intervals by solving an optimization problem with a finite time horizon, following a receding horizon approach. The optimization problem includes a model and constraints of the system to be controlled and forecasts of the exogenous information, such as user behaviour, outside temperature and solar radiance. When controlling a large cluster of flexibility sources however, solving the optimization problem centralised quickly becomes intractable~\cite{Mathieu}. To mitigate these scalability issues, distributed optimization techniques provide relief by decomposition of the master problem into sub-problems. Another approach (\textit{aggregate-and-dispatch}) gaining interest is to use a \textit{bulk} model of reduced order making the centralised optimization tractable by defining a setpoint for the entire cluster. Dissaggregation of the setpoint occurs through a heuristic dispatch strategy.

The mathematical performance of the aforementioned approaches however, is directly related to the fidelity of the model used in the optimization problem \cite{mathieu2013energy,Maasumy}. Obtaining and maintaining an accurate model, is a non-trivial task \cite{challengesMPC} where the cost of obtaining such a model can outweigh its financial benefits.

Given this context, model-free control solutions are considered a valuable alternative or addition to model-based control solutions \cite{ernst2009reinforcement}. The most popular model-free control paradigm is Reinforcement Learning (RL) \cite{sutton1998reinforcement}. For example when using Q-learning, an established form of RL, a control policy is learned by estimating a state-action value function, or Q-function based upon interactions with the system to be controlled. 
In \cite{WenZhen,RuelensBRLCluster,Giu}, RL has been been applied to the setting of residential demand response at the device level, whilst in \cite{RuelensBRLCluster} and~\cite{kara2012using}, RL has been used in an \textit{aggregate-and-dispatch} setting with a large cluster of TCLs.

In the literature, different approaches are presented that obtain an estimate of the Q-function. In this work, as in~\cite{RuelensBRLDevice}, batch RL~\cite{ernst2005tree} is used, where an estimate of the Q-function is obtained offline using a batch of historical tuples. A regression algorithm is used to generalize the estimate of the Q-function to unobserved state-action combinations. In \cite{RuelensBRLCluster}, extra-trees \cite{ernst2005tree} has been used as a regression algorithm in combination with hand-crafted features, furthermore perfect state information was assumed. This resulted in a low-dimensional state space, evading the curse of dimensionality. A next important step is to add automatic feature extraction as this enables to work with higher dimensional state representations. 
This in turn, allows to add historic observations to the state, following Bertsekas~\cite{bertsekas1996neuro}, as it can  compensate for  the partial observability of the state by extracting state-time features. Recent developments in the field of RL, and more specific deep reinforcement learning \cite{deepmind}, have demonstrated how by using a Convolutional Neural Network (CNN) automatic feature extraction can be obtained in a high-dimensional state space with local correlation. Inspired by these findings, this work applies deep reinforcement learning to the setting of \textit{aggregate-and-dispatch} with a high-dimensional state space, which includes past observations, allowing to extract state-time features, thus mitigating the effect of incomplete state information.

The remainder of this work is summarized as follows.
In Section~\ref{Sec:related_work} an overview of the related literature is provided and the contributions of this work are explained. 
Section~\ref{motivationes} sketches the main motivation behind this paper.
Following the approach presented in~\cite{RuelensBRLDevice}, in Section~\ref{Sec:problem_formulation} a Markov decision process formulation is provided.
In Section~\ref{ImplementationDDR}, the implementation of the controller is detailed, while Section~\ref{Sec:MABRL} presents a quantitative and qualitative assessment of its performance. 
Finally, Section~\ref{Sec:CDDR} outlines the conclusions and discusses future research.

\section{Related Work and Contribution}\label{Sec:related_work}
This section provides a non-exhaustive overview of related work regarding the control of heterogeneous clusters of TCLs, batch RL applied to load control and automated feature extraction. 
\subsection{Aggregate and dispatch}
As mentioned in Section \ref{Sec.introductionDDR}, two important challenges in model-based control of a cluster of TCLs are the dimensionality of the state-action space and obtaining a high-fidelity model. To mitigate these challenges, two important \textit{schools} can be identified (amongst others), i.e. that of distributed optimization and \textit{aggregate-and-dispatch}.
The general concept behind aggregate-and-dispatch techniques \cite{koch2011modeling,tracers} is to use a bulk model representing the dynamics of the cluster of TCLs instead of individual modelling at TCL level. Typically, this bulk model is of a reduced dimensionality, making a centralised MPC approach tractable. The subsequent set-points are dispatched at device level by using a simple heuristic, requiring little intelligence at device level.
For example in \cite{kara2012using}, the TCLs are clustered based upon their relative position within their dead-band, resulting in a state vector at cluster level containing the fraction of TCLs in each state bin. A linear state bin transition model describes the dynamics of this state vector, the dimensionality of which is independent of the number of TCLs in the cluster. This model is in turn used in an MPC, resulting in a control signal for each state bin. A simple heuristic is used to dispatch the control signals to individual control actions at device level. The results presented in \cite{mathieu2013energy} show that careful system identification is required and a generic \textit{tank} model is preferred for example for energy arbitrage. 
Moreover, in \cite{zhang2012aggregate} it is argued that the first order TLC model presented in \cite{mathieu2013energy} needs to be extended with a representative \textit{bulk} temperature that is not directly observable, necessitating a model-based state estimation such as a Kalman filter \cite{Mathieu}.  
Also in\cite{Callawayheterogeneous}, a low-order tank model is used as a bulk model to describe the flexibility of a large cluster of TCLs allowing for tractable stochastic optimization. A kindred approach is presented by Iacovella \textit{et al.} in \cite{tracers}. Here, a small set of representative TCLs are identified to model the dynamics of the entire cluster. Through this, the heterogeneity of cluster is accounted for and the auction-based dispatch dynamics can be included in the central optimization. This work is an extension of the work presented in \cite{Vandael2012TSA} where a tank model has been used for a cluster of electric vehicles, again in combination with an auction-based dispatch strategy. The main advantage of aggregate-and-dispatch is that it mitigates the curse of dimensionality allowing to hedge against uncertainty at a centralised level \cite{Callawayheterogeneous} and requires little and transparent intelligence at the level of a TCL, restricting the local cost. A system identification step at centralised level however, is still required.\\
\indent A different paradigm is that of distributed optimization \cite{Gatsis2012,DeRidderDD}, where the centralised optimization problem is decomposed over distributed agents who interact iteratively through \textit{virtual} prices which are the Lagrangian multipliers related to coupling constraints. For example in \cite{DeRidderDD}, distributed MPC through dual decomposition was presented as a means for energy arbitrage of a large cluster of TCLs subject to a coupling constraint related to an infrastructure limitation. Although distributed optimization techniques converge to a global optimum under sufficient conditions \cite{BoydCvx}, the technical implementation is not straightforward. This results from the need for a system identification step for each sub-problem (often at the level of a TCL) and the fact that on the order of ten iterations are necessary before convergence is obtained. 
Although the merits of distributed optimization are recognized, this work focuses on aggregate-and-dispatch techniques in the scope of residential flexibility, as the local intelligence at device level is simple and transparent.

\subsection{Reinforcement Learning for demand response}\label{Sec:RLreview}
As discussed in Section~\ref{Sec.introductionDDR}, RL is a model-free control technique whereby a control policy is learned from interactions with its environment. When integrated in an aggregate-and-dispatch approach, it allows to replace or assist a model-based controller \cite{Giu}. This paves the way for generic control solutions for residential demand response. 
When considering RL applied to aggregate-and-dispatch techniques, Kara \textit{et al.} applied Q-learning \cite{kara2012using} to the binning method presented in \cite{koch2011modeling}. In~\cite{RuelensBRLCluster}, Ruelens \textit{et al.} applied batch RL in the form of Fitted Q-Iteration (FQI) to a cluster of TCLs using an auction-based dispatch technique effectively learning the dynamic of the cluster, including uncertainty and effects of the dispatch strategy. A second implementation is presented in \cite{StijnBRL}, where FQI was used to obtain an accurate day-ahead consumption schedule for a cluster of electric vehicles. 
The focus was on finding a day-ahead schedule that results in the lowest electricity cost considering day-ahead and intra-day electricity prices.
Although the results demonstrated that RL is of interest for demand response, the state dimensionality was small and the features in the state were handcrafted, furthermore full observability was assumed. This is a limitation when considering a very heterogeneous cluster and partial observability.
In this setting, a richer state description is required, e.g. as in \cite{koch2011modeling} using a state bin distribution. Furthermore, following \cite{bertsekas1996neuro} the state vector needs to be extended to include previous observations as it allows to extract state-time features that can be representative for non-observable state information. This however, requires automatic high-dimensional feature extraction. 

\subsection{High-Dimensional Feature Extraction}
As mentioned in Section \ref{Sec.introductionDDR}, recent results show that deep approximation architectures such as CNNs can be used as a regression algorithm in RL applied to a problem with a high-dimensional state vector.
Artificial neural networks offer an attractive option for value function approximation. They can approximate general nonlinear functions, scale  to high-dimensional input spaces and can generalize to unseen inputs. Furthermore, deep network architectures stack multiple layers of representations and can be used with low level sensor inputs (e.g. image pixels) to learn multiple levels of abstraction and reduce the need to manually define features. 

Unfortunately, when used with sequential updating and correlated observations, as is typical in online reinforcement learning, neural networks can suffer from issues such as divergence of the estimates or catastrophic forgetting \cite{goodrichmitigating,boyan1995generalization}.  The FQI  algorithm~\cite{riedmiller2005neural} used in this paper, sidesteps this problem by relying on offline approximation of the value function using batch training of the function approximators. It has previously been applied to a range of control applications  \cite{hafner2007neural,gabel2011improved,kietzmann2009neuro}.  Additionally, FQI was successfully combined with deep architectures by Lange \textit{et al.} \cite{lange2010deep,lange2012autonomous}, who extended the algorithm using deep autoencoders to learn features from image pixel inputs. 
Another approach to combine neural networks with RL  was recently proposed by Mnih \textit{et al.} \cite{deepmind}. Here a database of transition samples is used with experience replay \cite{lin} to break correlations in the training set in online value function approximation \cite{lecun1998gradient}. As is the case in this paper, the proposed deep Q-network (DQN) algorithm uses CNN  architecture to map low level inputs to Q-values. Following this result a number of approaches combining reinforcement learning with CNNs have been proposed.  Guo \textit{et al.} \cite{NIPS2014_5421} combine a DQN agent with offline planning agents for sample generation. 
In~\cite{lillicrap2015continuous}, Lillicrap \textit{et al.}  use the DQN architecture in an actor-critic setting with continuous action spaces. Finally, Levine \textit{et al.} \cite{levine2015end} introduce a different approach using a CNN to represent policies in a policy search method. 
In this paper, we combine a CNN and a multilayer perceptron to approximate Q-values in the batch FQI setting.
It offers the following contributions:
\begin{itemize}
\item A merged Artificial Neural Network (ANN), comprising a CNN~\cite{deepmind} and a multilayer perceptron, tailored to a demand response setting, is used as a regression algorithm within FQI. 
To the best of our knowledge, this work is the first description of such a network to be used in combination with a batch reinforcement learning algorithm.
\item By presenting the CNN with a series of state-bin distributions, state-time features  that are relevant to learn near-optimal policies can be extracted. 
\item The resulting control strategy is evaluated on a simplified and qualitative test scenario comprising a heterogeneous cluster of TCLs with partial observability, exposed to a time varying price. 
The results demonstrate that the presented approach can be successfully applied to residential load control.
\end{itemize}

\section{Background and Motivation}
\label{motivationes}
As detailed in~\cite{zhang2012aggregate, reynders2014quality,vrettos2016}, the dynamics of a Thermostatically Controlled Load (TCL) is dominated by at least two time scales, a fast one (related to the operational air temperature) and a slow  one (related to the building mass).
A detailed description of the second-order dynamics of a TCL can be found in Section~\ref{sim_model}. 
This model describes the temperature dynamics of the indoor air and of the building envelope. 
Typically, only the operational air temperature is available from which all information needs to be extracted. In a model-based implementation all non-observable states are determined using a Kalman filter~\cite{vrettos2016}. However, before one can implement this filter, one first needs a calibrated model. This is typically a non-linear optimization problem as presented in~\cite{vrettos2016}. 
In a model-free approach, information regarding the non-observable states needs to be extracted from the last $N$ observations. This results in a severe extension of the state space. Driven by this challenge, this paper combines a batch RL technique with a convolutional neural network to make up for the partial observability of the problem. 

\section{Problem Formulation}\label{Sec:problem_formulation}
Before presenting the control approach in Section \ref{ImplementationDDR}, the decision-making process is formulated as a Markov Decision Process (MDP)~\cite{bertsekas1996neuro} following the procedure presented in \cite{RuelensBRLDevice}. An MDP is defined by its state space $X$, its action space $U$, and its transition function $f$:
\begin{equation}
\x_{k+1}=f(\x_{k},\action_{k},\w_{k}),
\end{equation}
which describes the dynamics from $\x_{k}\in X$ to $\x_{k+1}$, under the control action $\action_{k} \in U$, and subject to a random process  $\w_{k} \in W$, with probability distribution $p_{w}(\cdot,\x_{k})$. 
The cost $c_{k}$ accompanying each state transition is defined by:
\begin{equation}
c_{k}(\x_{k},\action_{k},\x_{k+1})=\rho(\x_{k},\action_{k},\w_{k}).
\end{equation} 
The objective is to find a control policy ${h:~X~\rightarrow~U}$ that minimizes the $T$-stage cost starting from state $\x_{1}$, denoted by $J^{h}(\x_{1})$:
\begin{equation}
J^{h}(\x_{1}) = \mathbb{E}\left(R^{h}(\x_{1},\w_{1},...,\w_{T})\right), 
\label{eq.J}
\end{equation}
with:
\begin{equation}
R^{h}(\x_{1},\w_{1},...,\w_{T}) = \sum_{k=1}^{T}{\rho(\x_{k},h(\x_{k}),\w_{k})}. 
\label{eq.rewardf}
\end{equation}

A convenient way to characterize the  policy $h$ is  by using  a state-action value function or Q-function:
\begin{align}
Q^{h}(\x,\action) = \underset{w\sim\probw}{\expected} \left[\rho(\x,\action,\w) +  J^{h}(f(\x,\action,\w)) \right].
\label{Qfunction}
\end{align}
The Q-function is the cumulative return starting from  state $\x$, taking action $\action$, and following $h$ thereafter.

The optimal Q-function corresponds the best Q-function that can be obtained by any policy:
\begin{equation}
Q^{*}(\x,\action) = \underset{h}{\text{min~}} Q^{h}(\x,\action).
\end{equation}
Starting from an optimal Q-function for every state-action pair, the optimal policy is calculated as follows:
\begin{equation}
h^{*}(\x)  \in \underset{\action \in U}{\text{arg min~}} Q^{*}(\x,\action),
\label{Qpolicy}
\end{equation}
where $Q^{*}$ satisfies the Bellman optimality equation~\cite{BellmanDP}:
\begin{align}
Q^{*}(\x,\action) = \underset{\w\sim\probw}{\expected}\left[\rho(\x,\mathbf{u},\w) + \underset{u' \in U}{\text{min~}} Q^{*}(f(\x,\action,\w),\action') \right].
\label{Qfunction}
\end{align}

\subsection{State description}
\label{Sec:state_description}

Following \cite{RuelensBRLDevice}, the state space $X$ consists of: time-dependent state information $\Xtime$, controllable state information $\Xphys$, and exogenous (uncontrollable) state information $\Xexo$. 

Since the problem of scheduling a TCL includes time dependence, i.e. the system dynamics are non-stationary, it is important to include at time-dependent state component to capture these patterns.
As in \cite{Giu}, the time-dependent state component $\Xtime$ contains information related to timing. In this work, this component contains the hour of the day:
\begin{equation}
\xtime \in \Xtime= \left\{1,\dots,24\right\}.
\label{eq.timestate}
\end{equation}
By adding a time-dependent state component to the state vector, the learning algorithm can capture the  behavioral patterns of the end users. The rationale is that most
consumer behavior tends to be repetitive and tends to follows
a diurnal pattern.

The controllable state information $\xphysk$ comprises the operational temperature $T_{k}^{i}$ of each TCL $i \in \mathcal{D}$: 
\begin{equation}
\underline{T}_{k}^{i}< T_{k}^{i} <\overline{T}_{k}^{i} \, 
\label{eq.physstate}
\end{equation} 
where $\underline{T}_{k}^{i}$ and $\overline{T}_{k}^{i}$ denote the lower and upper bound set by the end user. 

The exogenous state information $\xexok$ cannot be influenced by the control action $\action_k$, but has an impact on the physical dynamics.
In this work, the exogenous information comprises the outside temperature $\Tok$.
A forecast of the outside temperature $\Tohat$ \footnote{The notation $\hat{x}_{\mathrm{ex}}$ is used to indicate a forecast of the exogenous state information $x_{\mathrm{ex}}$.} is assumed available when calculating the Q-function, as detailed in Section~\ref{Sec:FQI}.

The observable state vector $\xobsk$ of the cluster is defined as:
\begin{equation}
\xobsk =\left(x_{\mathrm{t}, k},T_{k}^{1},\ldots,T_{k}^{|\mathcal{D}|},\Tok\right). 
\label{eq.stateDef}
\end{equation}

\subsection{Backup controller and physical realisation}
\label{subsection.backup_controller}
The control action for each TCL is a binary value indicating if the TCL needs to be switched ON of OFF:
\begin{equation}
u_{k}^{i} \in \left\{0,1\right\}. 
\label{eq.contr_space}
\end{equation}
Similar as in~\cite{koch2011modeling} and~\cite{RuelensBRLDevice}, each TCL is equipped with a backup controller, acting as a filter for the control action resulting from the policy $h$. 
At each time step $k$, the function  $B$ maps the requested control action $u_{k}^{i}$ of device $i$ to a physical control action $\uphysk$, depending on its indoor air temperature $T_{k}^{i}$: 
\begin{equation}
\uphysk = B(T_{k}^{i},u_{k}^{i},\bm{\theta}^{i}),
\label{Eq:backup_controller_1}
\end{equation} 
where $\boldsymbol{\theta}^{i}$ contains the minimum and maximum temperature boundaries, $\underline{T}_{k}^{i}$  and $\overline{T}_{k}^{i}$ set by the end user and $B \left( \cdot \right)$ is defined as:
\begin{align}
B(T_{k}^{i},u_{k}^{i},\theta^{i}) = \left\{\begin{matrix}
1&\text{if }& {T_{k}^{i} \leq}\underline{T}_{k}^{i} \quad \quad \quad \quad \;\;\, \\ 
u_k^{i}&\text{if }& {\underline{T}_{k}^{i} \leq } {T_{k}^{i} \leq}\overline{T}_{k}^{i}.\quad \quad \\ 
0 &\text{if }& {T_{k}^{i} >}\overline{T}_{k}^{i} \quad \quad \quad \quad \;\;\,
\end{matrix}\right.
\label{Eq:backup_controller_2}
\end{align}
The backup controller guarantees the comfort settings of the end user by overruling the requested control action $u_k^{i}$ when the comfort constraints of the end user are violated. For example, if the temperature of TCL $i$ drops below  $\underline{T}_{k}^{i}$ the backup controller will  activate the TCL, independent of the requested control action, resulting in $\uphysk$, which is needed to calculate the cost~(\ref{eq.rewardarbitrage}).

\subsection{Cost function}
Different objectives are considered in the literature when controlling a large cluster of TCLs, for example, tracking a balancing signal \cite{koch2011modeling} or energy arbitrage \cite{mathieu2013energy}. 
In this work, we consider energy arbitrage, where TCLs can react to an external price vector $\bm{ \lambda}$. 
The cost function $\rho$  is defined as:
\begin{equation}
\rho \left(\xobsk, \uphysI,\ldots, \uphysD, \lambda_k \right) = \Delta t \lambda_{k} \sum_{i=1}^{|\mathcal{D}|}{\uphysk P^{i}}\; ,
\label{eq.rewardarbitrage}
\end{equation}
where $P^{i}$ is the average power consumption of the $i$th TCL during time interval $\Delta t$ and $\lambda_{k}$ is the electricity price during time step $k$.

\begin{figure*}[t!]
\centering{\includegraphics[width = 12 cm]{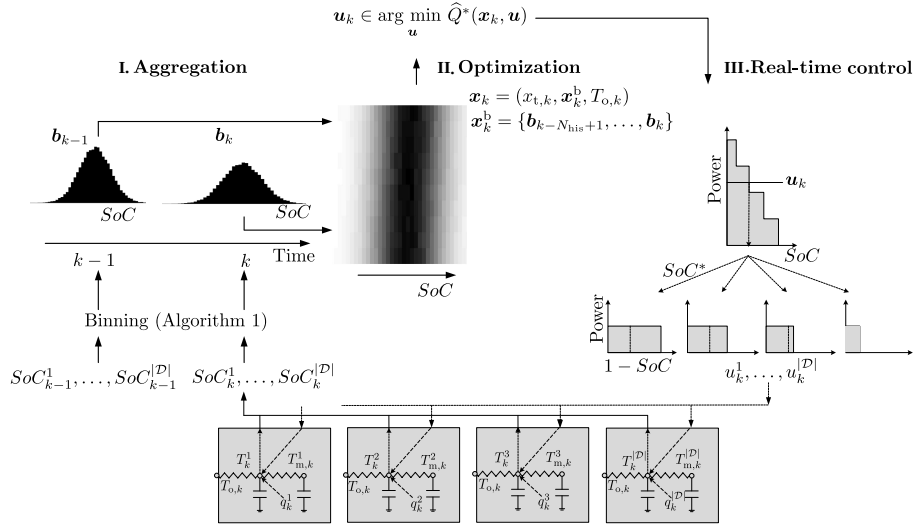}}
\caption{
\small Three steps are identified, i.e. aggregation, optimization and real-time control. The models indicated at the bottom only provide access to the room temperature $\Ta$, whilst the temperature of the building mass $\Tm$ is hidden.}
\label{fig.OverViewDTSA}
\end{figure*}

\section{Implementation}
\label{ImplementationDDR}

In this section the implementation details of the  presented  controller  are described. Similar as in~\cite{koch2011modeling} and~\cite{Vandael2012TSA} a three-step approach is used (Fig.~\ref{fig.OverViewDTSA}). 

\subsection{Step 1: Aggregation}
\label{s.stepOne}
In the first \textit{aggregation} step, an aggregated state representation is created from $\Nh$ historical observations $\xobs_{k-\Nh+1},\ldots,\xobsk$. 
Each observation $\xobsk$ is processed similar as in \cite{koch2011modeling}, i.e. each TCL in the state vector $\xobsk$ is binned according to its state of charge ($SoC^{i}_{k}$) in the binning vector $\bvet_k$  with support points $\bs$. 
The vector $\bs$ contains $\Nb$ equidistant points between the minimum and maximum state of charge of the cluster.
For each TCL in the cluster, Algorithm ~\ref{binningAlgorithm} calculates the corresponding $SoC^{i}_{k}$ (line 3) and allocates this $SoC^{i}_{k}$ within the corresponding state of charge interval (line 4 and 5), indexed by $j \in  \{1,\ldots,\Nb\}$.

\begin{algorithm}[t]
\caption{Calculate the binning vector $\bvet_{k}$.}
\label{binningAlgorithm}
\begin{algorithmic}[1] 
\algsetup{linenosize=\tiny}
\renewcommand{\algorithmicrequire}{\textbf{Input:}}
\REQUIRE $ \xobsk,\bs$\\
\STATE let $\bvet_{k}$ be zeros everywhere on $\mathbb{N}^{|\bs|}$\\
\FOR {$i=1,\ldots,|\mathcal{D}|$}
\STATE $\SoC_{k}^{i}=\frac{T_{k}^{i}-\underline{T}_{k}^{i}}
{\overline{T}_{k}^{i}-\underline{T}_{k}^{i}}$
\STATE $j^{*}$ = arg max$_{j} \bsj$\\
\STATE ~~~~~~$\mathrm{s.t.}~ \bsj\leq \SoC_{k}^{i}$
\STATE $b_{k,j^{*}}\leftarrow b_{k,j^{*}}+1$
\ENDFOR
\ENSURE $\bvet_{k}$
\end{algorithmic}
\end{algorithm}

In a second step, the binning vectors of subsequent time steps are concatenated, resulting in $\xbk \in \mathbb{R}^{\Nb \times \Nh} $:
\begin{equation}
\xbk = \{\bvet_{k-\Nh+1},\ldots,\bvet_{k}\},
\end{equation}
where $\Nh$ denotes the number of historical time steps included in $\xbk$.
As a result, the final state vector is defined as: 
\begin{equation}
\x_{k} =\left(\xtimek,\xbk,\Tok\right). 
\label{eq.stateDef}
\end{equation}
\subsection{Step 2: Batch Reinforcement Learning}
\label{s.BRL}
In the second step, a control action for the entire cluster is determined following (\ref{Qpolicy}). In this work, FQI is used to obtain an approximation $\widehat{Q^{*}}$ of the state-action value function $Q^{*}$ from a batch of four tuples $\mathcal{F}$, as detailed in \cite{ernst2005tree}:
\begin{equation}
\mathcal{F} =\left\{(\x_{l},u_{l},\x_{l}',c_{l}),\, l = 1,...,\#\mathcal{F}\right\},
\label{eq.tuples}
\end{equation}


\subsubsection{Fitted Q-Iteration}
\label{Sec:FQI}
Building upon recent results \cite{RuelensBRLDevice,Giu}, FQI is used to obtain $\widehat{Q^{*}}(\x,u)$. 
The cost function is assumed known (\ref{eq.rewardarbitrage}) and the resulting actions of the backup controller can be measured. As a consequence,  Algorithm~\ref{forecastedFQI} uses tuples of the form
$\left(\x_{l},u_{l},\x_{l}',\uphysl \right)$.
Here $\x_{l}'$ denotes the successor state to $\x_l$. To leverage the availability of forecasts, in
Algorithm~\ref{forecastedFQI} the observed exogenous information (outside temperature) in $\xexol'$ is replaced by its forecasted value $\xhatexol'$ (line 5 in Algorithm~\ref{forecastedFQI}).
In step 6, a neural network is used $|U|$ times to determine the minimum value of  the current approximation of the Q-function $\widehat{Q}_{N-1} (\bm{\hat{x}}_{l}^{\prime},.)$.
In step 8, the neural network, used in step 6, is trained using all tuples $(\bm{x}_{l},u_{l})$ as input and all Q-values $Q_{N,l}$ as output data.

In our previous work~\cite{RuelensBRLCluster,StijnBRL}, an ensemble of extremely randomized trees~\cite{ernst2005tree} was used as a regression algorithm to  estimate the Q-function.
Given the high dimensionality of the state $(\xb_k \in \mathbb{R}^{\Nb \times \Nh}$) and given that state-time features are expected to have strong local correlations, this work proposes an artificial neural network with a convolutional component.

\begin{algorithm}[t]
\caption{Fitted Q-iteration using a convolutional neural network to extract state-time features.}
\label{forecastedFQI}
\begin{algorithmic}[1] 
\algsetup{linenosize=\tiny}
\renewcommand{\algorithmicrequire}{\textbf{Input:}}
\REQUIRE $\mathcal{F}=\{\x_{l}, u_l, \x_{l}', \uphysl\}_{l=1}^{\#\mathcal{F}}, \Xhatexo= \left\{\xhatexok\right\}_{k=1}^{T}, \bm{\lambda}$ \\
\STATE let $\widehat{Q}_{0}$ be zero everywhere on $X$ $\times$ $U$
\FOR {$N=1,\ldots,T$}
\FOR {$l = 1,\ldots,\#\mathcal{F}$}
\STATE $~~c_{l} \leftarrow \rho (\x_{l}, \uphysl, \bm{\lambda})$
\STATE $~~\hat{\x}'_{l} \leftarrow ({x}_{\mathrm{t},l}',\xblprime,\xhatexol')$
\STATE $~~Q_{N,l}\leftarrow c_{l} +\underset{u \in U}{\text{min~}}\widehat{Q}_{N-1}(\hat{\x}_{l}',u)  $  
\ENDFOR
\STATE use the convolutional neural network in Fig.~\ref{fig.DDRArchitecture} to obtain $\widehat{Q}_{N}$ from $\mathcal{T} = \left\{\left((\x_{l},u_{l}),Q_{N,l}\right),l =1,\ldots,\#\mathcal{F}\right\}$ 
\ENDFOR
\ENSURE $\widehat{Q}^{*}=\widehat{Q}_{N}$
\end{algorithmic}
\end{algorithm}

\begin{figure*}[t!]
\centering{\includegraphics[width = 13 cm]{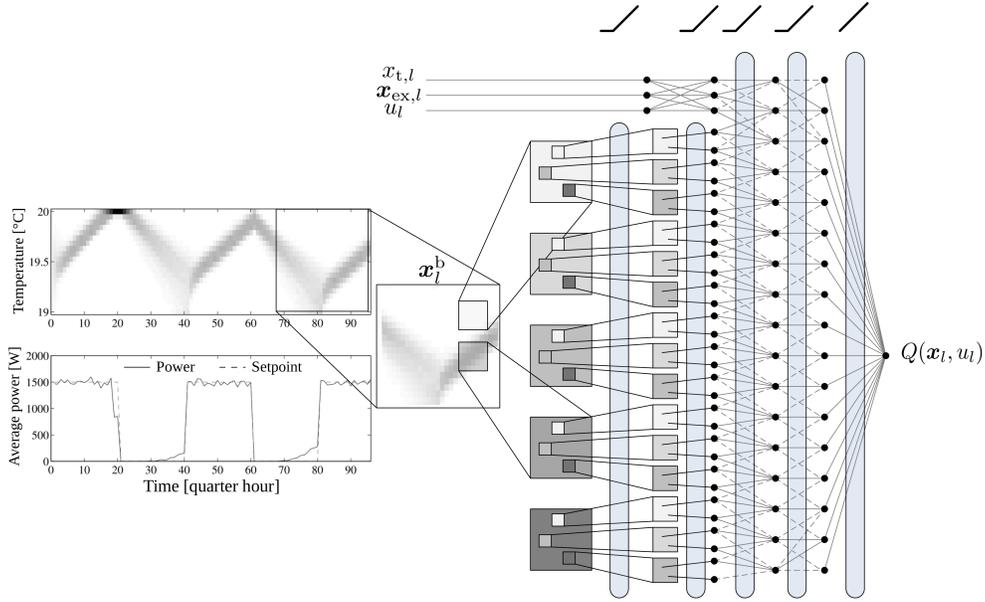}}
\caption{\small Overview of the regression step as used in the fitted Q-iteration implementation (line 8 in Algorithm~\ref{forecastedFQI}). The controllable state information, represented in the form of a matrix,  goes into two convolutional layers that identify state-time features. The other parts of the state, i.e. time-dependent and exogeneous state information together with the control action go through a dense neural network where also features can be extracted. Finally both layers are merged followed by two fully connected layers.}
\label{fig.DDRArchitecture}
\end{figure*}
\subsubsection{Regression Algorithm}\label{Sec:Regression}
The parametrization of the Q-function is given by a neural network architecture consisting of two  subcomponents: a convolutional component and a standard multi-layer perceptron.  The full architecture is shown in Fig.~\ref{fig.DDRArchitecture}. The network takes as input a state action pair $(\x,u)$ and returns an approximated Q-value $\widehat{Q}^*(\x,u)$. 
 The  inputs of the neural network are split into two parts.
The first part contains a $\Nb \times \Nh$ grid corresponding to the binned state representations $\xb$ and the second part contains the time-dependent state information $\xtime$, the exogenous state information $\xexo$ and the action $u_{k}$.
The binned state representation $\xb$ is  processed using  a CNN. CNNs process inputs structured as a 2-dimensional grid (e.g. images, video) by convolving the input grid with multiple linear filters with learned weights. In this way, CNNs can learn to detect spatial features in the local structure of the input grid.  A convolutional layer consists of multiple filters $W^k$, each giving rise to an output \emph{feature map}. The feature map $h^k$ corresponding to the $k$th filter weight matrix $W^k$ is obtained by:
\begin{equation}
 h^k_{ij} = \sigma(W^k * x)_{ij} + b^k,
\end{equation}
where $*$ represents a 2d convolution operation, $x$ are the layer inputs, $b^k$ is a bias term and $\sigma$ is a nonlinear activation function. Multiple layers can be stacked to obtain a deep architecture. Convolutional  layers are often alternated with pooling layers that downsample their inputs to introduce an amount of translation invariance into the network. Convolutional and pooling layers are followed by fully connected layers that combine all the feature maps produced by the convolutional part and produce the final network outputs.
The CNN processes the binned $\xb \in \mathbb{R}^{\Nb\times \Nh}$ with 1 dimension of the input grid corresponding to the $\Nb$ bins and the other dimension representing observations at $\Nh$ previous time steps. Time and state dimensions are treated equally and 2d convolution operations are applied over both these dimensions. This results in the identification of spatio-temporal features that identify local structure in the state information and history. This enables the network to identify features corresponding to  events that occur over multiple time steps. These features are then used as input by higher network layers.
The time-dependent state information $\xtimek$, exogenous input values $\xexok$ and actions $u_k$ are fed into a separate fully-connected feedforward architecture. This multi-layer perceptron first maps the inputs to an intermediate hidden representation. 
This hidden representation is then combined with the CNN output and both networks are merged into fully connected layers. A final linear output layer maps the combined hidden features of both networks to the predicted Q-value of the input state-action pair. This 2-stream network structure that combines different sources of information is similar to the organisation used in supervised problems with multimodal inputs~\cite{ngiam2011multimodal}.

\subsection{Step 3: Real-time control}
\label{s.realtime}
In the third step, a control action for the entire cluster $u_{k}$ is selected using an $\varepsilon$-greedy strategy, where
the exploration probability is decreased on a daily basis according to a harmonic sequence~\cite{sutton1998reinforcement}. 
Following \cite{Vandael2012TSA}, $u_{k}$ is dispatched over the different TCLs using an auction-based multi-agent system. 
In this market, each TCL is represented by a bid function $\bidfunction$,
which defines the power consumed versus a heuristic $\prio$, resulting in the following expression for each TCL $i$:
\begin{equation}
\bidfunction(\prio) =
\left\{\begin{matrix}
P^{i}  & \text{if}  &0 < \prio \leq \cornerprio\\ 
0               &  \text{if} &   \prio >\cornerprio
\end{matrix}\right. \; ,
\end{equation}
where $\cornerprio$ is the corner priority.
The corner priority indicates the wish (priority) for consuming at a certain power rating $P^{i}$. 
The closer the state of charge  drops to zero, the more urgent its scheduling (high priority), the closer to  $1$, the lower the scheduling priority.
The corner priority of the $i$th TCL  is given by $\cornerprio = 1 - \SoC^{i}$, 
where $SoC^{i}$ is the state of charge of TCL $i$. 
At the aggregated level, a clearing process is used to translate the aggregated control action $u_{k}$ to a clearing priority $\priostar$ : 
\begin{equation}
\priostar ~= ~\underset{\prio}{\mathrm{arg~min}} \left|\sum_{i=1}^{|\mathcal{D}|}{\bidfunction(\prio})-u_{k}\right|.
\label{eq.clearing}
\end{equation}
Note, in~(\ref{eq.clearing}) the clearing priority $\priostar$ is found by matching the aggregated control action $u_{k}$  in the aggregated bid function of the cluster.
This clearing priority $\priostar$  is sent back to the different TCLs, who start consuming according $u_{k}^{i}=\bidfunction(\priostar)$.

\section{Results}
\label{Sec:MABRL}
In order to evaluate the functionality of the controller presented in Section \ref{ImplementationDDR}, a set of numerical experiments were performed on a qualitative scenario, the results of which are presented here.
The simulation scenario comprises a cluster of 400 TCLs exposed to a dynamic energy price~\cite{belpex}. The thermal inertia of each TCL is leveraged as a source of flexibility allowing the electric demand to be shifted towards moments when the energy price is low. A backup controller (\ref{Eq:backup_controller_2}) deployed at each TCL safeguards the comfort constraints.
In the following simulation experiments, we define a control period of 1 hour and an optimization horizon of 24 control periods. At the start of each optimization horizon Algorithm 2 is used to compute a control policy for the next 24 control periods. This control policy is updated every 24 hours using a new price profile and forecast of the outside temperature.
During the day, an $\varepsilon$-greedy exploration strategy is used to interact with the
environment and to collect new transitions that are added
systematically to the given batch.
Since more interactions result in a better coverage of the state-action space, the exploration probability ${\varepsilon}_d$ is decreased on a daily basis, 
according to the  harmonic sequence $1/d^{n}$, where $n$ is set to 0.7 and $d$ denotes the current day.

\begin{figure}[t!]
\centering{\includegraphics[width=1.0\columnwidth]{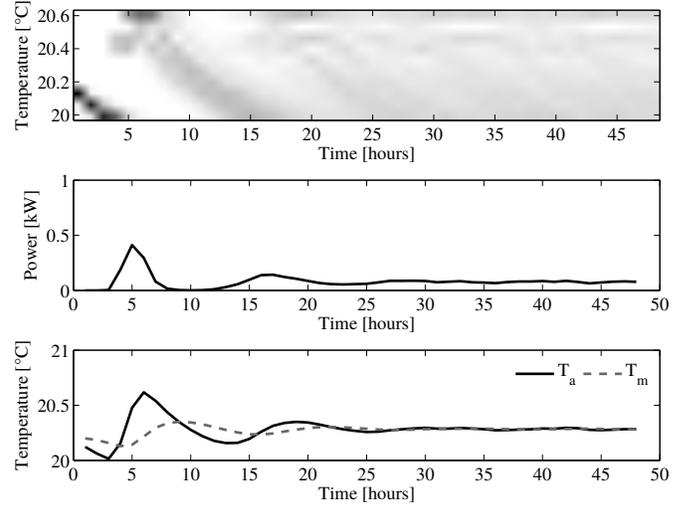}}
\caption{\small Top graph, evolution of the distribution of the TCL population over time. Middle graph, the average power evolving over time, after about 30 hours full decoherence is observed. Lower graph, the average observable and non-observable temperature as a function of time.}
\label{fig.TCLs}
\end{figure}
\subsection{Simulation model}
\label{sim_model}
Following \cite{tracers,zhang2012aggregate}, a second-order model has been used to describe the dynamics of each building as illustrated in Fig.~\ref{fig.OverViewDTSA}: 
\begin{equation}
\begin{matrix}
\dot{\Ta^{i}} & = &\frac{1}{\Ca^{i}} \left( \To-\Ta^{i}\right)&+ \frac{1}{\Cm^{i}}\left(\Tm^{i}-\Ta^{i}\right)+P^{i}u^{i}+q^{i},\\
\dot{\Tm^{i}} & = &\frac{1}{\Cm^{i}} \left( \Ta^{i}-\Tm^{i}\right),&\label{eq.secondOrder}
\end{matrix}
\end{equation}
where $\Ta^{i}$  is the measured indoor air temperature and $\Tm^{i}$ is the not observable building mass temperature. 
For each TCL in the simulation, the values $1/\Ca^{i}$ and $1/\Cm^{i}$ are selected random from a normal distributions $\mathcal{N}(0.004,0.0008)$ and $\mathcal{N}(0.2,0.004)$ respectively. The internal heating $q$ is sampled from $\mathcal{N}(0,0.01)$ for each time step. 
The power $P^{i}$ is set to 0.5 kW for each TCLs and the minimum and maximum temperatures are set at 20 and 22$^{\circ}$C for each TCL.
To illustrate the effect of the heterogeneity of parameters, Fig.~\ref{fig.TCLs} depicts the temperature dynamics of 1000 TCLs where a backup-back controller \cite{Giu} is used. 
The top graph in Fig.~\ref{fig.TCLs} shows the evolution of the temperature distribution, initially all TCLs have the same state, however, after one day de-phasing has occurred which is a direct measure for the heterogeneity of the cluster. The middle graph shows the aggregated power as a function of time. The initial coherence, resulting in strongly correlated consumption, is gone after around 30 hours. The lower graph shows the average values of $\Ta$ and $\Tm$ respectively.
As discussed in Section \ref{ImplementationDDR}, only $\Ta$, is assumed available, whilst features representing $\Tm$ are inferred from past measurements of $\Ta$ by the convolutional section in the regression algorithm. The coefficients of the model were chosen as such that $\Cm$ has a small but significant effect on the objective.

\subsection{Theoretical benchmark}
An optimal solution of the considered control problem is found by using a mathematical solver~\cite{CPLEX}. The objective of the benchmark is to minimize the electricity cost of the cluster:
\begin{equation}
\mathrm{min} \sum_{i=1}^{|\mathcal{D}|} \sum_{t=1}^{T} \lambda_{t} \uphyst P^{i}
\end{equation}
subject to the second-order models~(\ref{eq.secondOrder}) and comfort constraints~(\ref{Eq:backup_controller_2}) of the individual TCLs in the cluster. 
Note, the benchmark optimization has \textit{perfect} information  about the model and has \textit{full} access to the temperature of the building mass,  resulting in a mathematical optimal result. This result can be seen as a lower limit on the cost, indicating how far from the optimum the controller is.

To demonstrate the impact of ignoring the non-observable state $\Tm$ on the objective, the theoretical benchmark optimization was performed for coefficients corresponding to the mean of previous normal distributions. 
Ignoring the non-observable state $\Tm$ resulted in a cost increase of 2.5\%. 

\subsection{Deep regression architecture}
This subsection describes the exact architecture of the neural network depicted in Fig. 2 used during the simulations. A fragment of the Python code of the neural network can be found in the appendix section. The input of the CNN is provided by  $\xb \in \mathbb{R}^{\Nb \times \Nh} $.
In the simulations, the number of bins $\Nb$ is set to 28 and the number of previous time steps 
$\Nh$ to 28, resulting in a $28\times28$ grid.
The first hidden layer convolves four $7\times7$ filters with stride 1 with the input $\xb$ and applies a rectifier nonlinearity (ReLu).
The second hidden layer convolves eight $5\times5$ filters with stride 1, again followed by a rectifier nonlinearity. The convolutional layers are followed by a single fully connected layer mapping the feature maps to $32$ hidden nodes.
 The time-dependent state component $\xtimek$, exogenous state component $\xexok$ and action $u_k$ are processed using a single, fully connected  hidden layer of $16$ units. The combined output of the CNN and feedforward network are processed using two fully connected layers, each consisting of $24$ units. All layers used ReLu activations and no pooling layers were used. The final hidden representation is mapped to a single output using a fully connected linear output layer with a single hidden output.  
The network was trained using the RMSProp algorithm~\cite{RMSProp} with minibatches of size 16.

\subsection{Results}
\begin{figure}[t!]
\centering{\includegraphics[width=1.0\columnwidth]{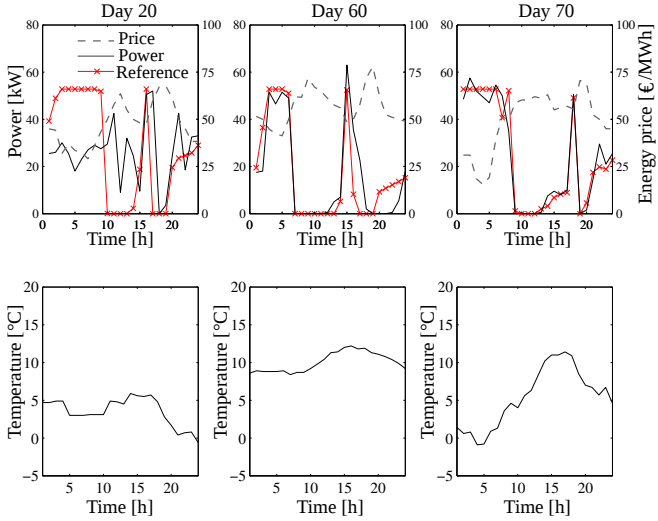}}
\caption{\small Illustration of the learning process, the black lines in top row depicts the power profiles obtained with the approach presented in this paper after 20, 60 and 70 days respectively, the red lines indicate profiles corresponding to a benchmark solution. Depicted with the dashed lines are the corresponding price profiles, whilst the lower graphs depict the corresponding outside temperature.}
\label{fig.OV}
\end{figure}

The simulations span a period of 80 days, each simulation taking about 16 hours\footnote{Intel\textsuperscript{R} Core \textsuperscript{TM} i5 2.5 GHz, 8192 MB RAM}. 
During the last eight days, the exploration probability was set to zero, resulting in a completely greedy policy according to (\ref{Qpolicy}).
In Fig.~\ref{fig.OV} one can see a selection of the results of the presented approach after different number of days for different outside temperatures. The number of days are indicated in the titles of the top row, i.e. after 20, 60 and 70 days. The bottom row depicts the corresponding outside temperatures. Added in the graph are the results of a benchmark optimization as discussed above. It is observed that the results obtained after 60 and 70 days are close to optimal and this for different outside temperatures.

\begin{figure}[t!]
\centering{\includegraphics[width=9cm]{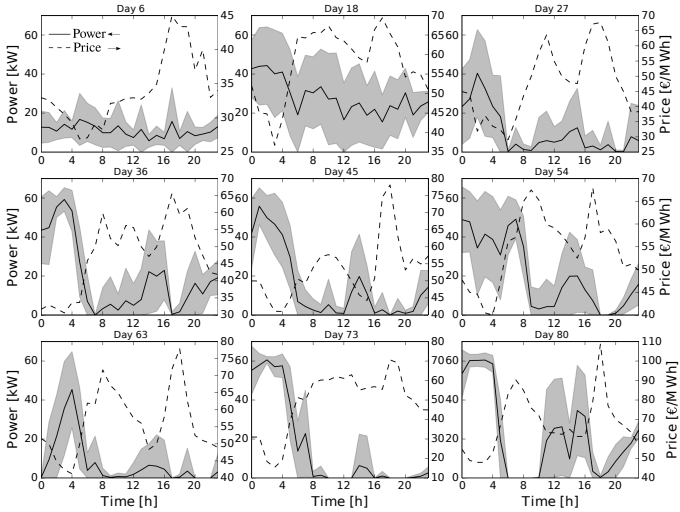}}
\caption{\small  Average power consumption  of the cluster (black line) surrounded by a gray envelope containing   95\% of the  power consumption profiles (of six simulation runs) for different days during the learning process (left y-axis). Daily price profiles  (dashed line, right y-axis).}
\label{extra_plotjes}
\end{figure}

Since a random exploration term is used, the experiments were repeated 6 times.  The results of these 6 simulation runs can be seen in Fig.~\ref{extra_plotjes}.  This figure depicts the power consumption profiles and corresponding electricity prices of the cluster for different days during the learning process. The last two subplots in the bottom row of the figure correspond to the last eight days obtained with a pure greedy policy.

\begin{figure}[t!]
\centering{\includegraphics[width=9cm]{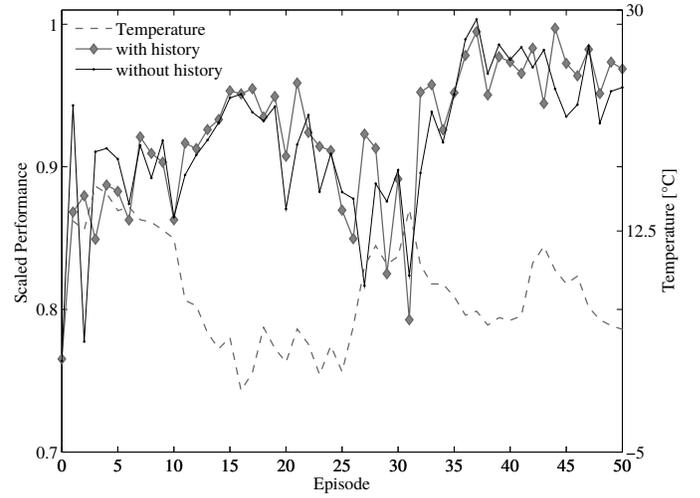}}
\caption{\small Illustration of the learning process, depicted is the scaled performance (averaged over six runs) with and without including the state-time features. Also depicted is the average outside temperature.}
\label{fig.convergence}
\end{figure}

This is presented more quantitatively in Fig.~\ref{fig.convergence}, where the scaled performance is
depicted. The scaled performance is defined as the daily cost  using our approach (Algorithm 2)  scaled with the daily cost  obtained using the theoretical benchmark. 
The results in Fig.~\ref{fig.convergence} are obtained by averaging the scaled performance over 6 statistical runs. 
It is observed that it takes on the order of 30 days before the control policy converges to a scaled performance of 0.95, after which its performance remains  stable.
Note that in Fig.~\ref{fig.convergence}, a scaled performance of one corresponds to the solution of the theoretical benchmark. 
\begin{table}
	\centering
		\caption{Overview of simulation results.\\
		A: without state-time features. B: with state-time features}
		\begin{tabular}{c| c   c   c  c  c c}
			\toprule
			  Experiment &I & II & III& IV& V&VI \\
			\midrule  							
						A & 1.0176 & 1.0255 & 0.9924 & 1.0043 & 0.9958 & 1.0157\\
						B & 0.9774 & 1.009 & 0.9922 & 0.9917 & 0.9902 & 0.9873\\
			\midrule 
						 $t$-test & $p$ &  3.2\% &  &   &  & \\
						\bottomrule

		\end{tabular}
    	\label{tab:Ressims}
\end{table}
\subsection{State-time features}
To identify the contribution of taking into account the history of observations into the state, a set of numerical experiments (spanning 80 days) has been conducted, the results of which are presented in Table \ref{tab:Ressims}. Six numerical experiments have been performed where the history of observations is added as discussed in Section~\ref{ImplementationDDR}. Similar, six numerical experiments have been conducted where the history of observations was omitted. In order to evaluate the contribution of the information present in the past observations, the network architecture has been left unchanged. The state however, has been constructed by copying the last observation $b_{k}$, $\Nh$ times. 
Table \ref{tab:Ressims} presents the scaled cumulated cost for both sets of simulations, i.e. with and without taking into account the history of observations. The cumulated cost is calculated for the last 30 days to make the results less sensitive for the effects of exploration. For clarity, the results are normalised with the mean cumulated cost over the twelve experiments. As the expected difference is on the order of one to two percent, a two-sample t-test has been conducted indicating that with almost 97\% probability the results originate from distributions with a different mean. Adding the history of observations to the state, reduces the average cost by approximately 1.2\%.  

\section{Conclusions}\label{Sec:CDDR}
Driven by recent successes in the field of deep learning~\cite{deepmind}, this work has demonstrated how a neural network, containing convolutional layers, can be used within fitted Q-iteration in a realistic demand response setting with partial observability.
 By enriching the state with a sequence of past observations, the convolutional layers used to obtain an approximation of the Q-function were able to extract state-time features that mitigate the issue of partial observability. The approach has been evaluated in a qualitative simulation, comprising a heterogeneous cluster of thermostatically controlled loads, that only share their operational air temperature, whilst their \textit{envelope} temperature remains hidden. 
The simulation experiments have demonstrated that our approach was able to obtain near-optimal results and that the regression algorithm was able to benefit from the sequence of past observations. 

Future work will be oriented towards  the application of high-fidelity building models, which we already started exploring in~\cite{aertgeerts2015agent}, and testing the performance of the proposed approach for other objectives such as tracking a reference profile.
In terms of research, other regression techniques, such as long short-term memory networks, will be investigated~\cite{LSTMs}.

\section*{Appendix: Convolutional Neural Network Architecture}

The following fragment of Python code shows the construction of the neural network (Fig.~\ref{fig.DDRArchitecture}) used during the simulations (Section~\ref{Sec:MABRL}). 
The implementation of the neural network was done using Keras~\cite{keras} and Theano~\cite{theano}.\\

\begin{python}
from keras.optimizers import RMSprop
from keras.models import Sequential
from keras.layers.core import (Dense, Activation, Merge, 
	             Flatten, Reshape, Convolution2D)

width1 = 7 # width first filter
CNN = Sequential()
CNN.add(Dense(28*28,28*28))
CNN.add(Reshape(1, 28, 28))
CNN.add(Convolution2D(4,1,width1,width1,
                                 border_mode='valid')) 
CNN.add(Activation('relu'))

width2 = 5 # width second filter
CNN.add(Convolution2D(8, 4, width2, width2))
CNN.add(Activation('relu'))
CNN.add(Flatten())
scaledGraph = 28-width1+1-width2+1
CNN.add(Dense(8*scaledGraph*scaledGraph, 32))
CNN.add(Activation('relu'))

model = Sequential()
model.add(Merge([CNN,Dense(2,16)], mode='concat'))
model.add(Dense(48,24))
model.add(Activation('relu'))
model.add(Dense(24,24))
model.add(Activation('relu'))        
model.add(Dense(24, 1))

# RMSProp optimizer [44]
Rpr = RMSprop(lr=0.001,rho=0.9,epsilon=1e-6)
model.compile(loss='mean_squared_error',optimizer=Rpr) 
\end{python}


\section{Acknowledgments}
This research is supported  by  IWT-SBO-SMILE-IT, funded by the Flemish Agency for Innovation  through Science IWT  through Science and Technology, promoting Strategic Basic Research.

\ifCLASSOPTIONcaptionsoff
  \newpage
\fi

\bibliographystyle{IEEEtran}  
\bibliography{references} 

\end{document}